\documentclass[conference]{IEEEtran}
\IEEEoverridecommandlockouts
\usepackage{amsmath,amssymb,amsfonts}
\usepackage{algorithmic}
\usepackage{graphicx}
\usepackage{textcomp}
\usepackage{xcolor}

\PassOptionsToPackage{sort&compress,numbers}{natbib}
\usepackage{natbib}

\usepackage{color}
\usepackage{soul}
\usepackage{multirow}
\usepackage[dvipsnames]{xcolor}
\usepackage{wrapfig}

\definecolor{darkred}{rgb}{0.7,0.1,0.1}
\definecolor{darkgreen}{rgb}{0.1,0.7,0.1}
\definecolor{cyan}{rgb}{0.7,0.0,0.7}
\definecolor{dblue}{rgb}{0.2,0.2,0.8}
\definecolor{maroon}{rgb}{0.76,.13,.28}
\definecolor{burntorange}{rgb}{0.81,.33,0}
\definecolor{tealblue}{rgb}{0.212,0.459, 0.533}
\definecolor{myyellow}{rgb}{0.8627451 , 0.75294118, 0.20784314]}

\definecolor{mypink}{rgb}{0.93359375, 0.62109375, 0.83984375}

\definecolor{pp}{rgb}{0.43921569, 0.18823529, 0.62745098}
\definecolor{rr}{rgb}{0.5254902 , 0.00784314, 0.12941176}
\definecolor{bb}{rgb}{0.09019608, 0.23529412, 0.37647059}
\definecolor{yy}{rgb}{0.49803922, 0.3372549 , 0.0}
\definecolor{gg}{rgb}{0.02352941, 0.3372549 , 0.17647059}
\definecolor{mybrown}{rgb}{0.87058824, 0.56078431, 0.01960784}
\definecolor{myblue}{rgb}{0.3372549 , 0.70588235, 0.91372549}
\definecolor{mypurple}{rgb}{0.8, 0.47058824, 0.7372549 }
\definecolor{myorange}{rgb}{0.835, 0.368, 0}
\definecolor{mygreen}{rgb}{0.00784314, 0.61960784, 0.45098039}
\definecolor{mygt}{rgb}{0.0078125 , 0.57421875, 0.40625}
\definecolor{mysp}{rgb}{0.84765625, 0.515625  , 0.0234375}
\definecolor{mycitecolor}{rgb}{0,0.08,0.45}
\definecolor{mygr}{rgb}{0.9607,0.9607,0.9607}
\definecolor{myoo}{rgb}{0.992,0.9176,0.9019}

\definecolor{myrr}{HTML}{AE031A}
\definecolor{mybb}{HTML}{0155B3}

\usepackage{amsmath,amsfonts,bm}

\def\1{\bm{1}}

\def\sp{space}

\def\vc{{\bm{c}}}

\def\vg{{\bm{g}}}

\def\vl{{\bm{l}}}

\def\vp{{\bm{p}}}
\def\vq{{\bm{q}}}

\def\mI{{\bm{I}}}

\def\mL{{\bm{L}}}

\def\mS{{\bm{S}}}

\DeclareMathAlphabet{\mathsfit}{\encodingdefault}{\sfdefault}{m}{sl}
\SetMathAlphabet{\mathsfit}{bold}{\encodingdefault}{\sfdefault}{bx}{n}

\def\gB{{\mathcal{B}}}

\def\gG{{\mathcal{G}}}

\def\gM{{\mathcal{M}}}

\def\gS{{\mathcal{S}}}
\def\gT{{\mathcal{T}}}

\def\sR{{\mathbb{R}}}
\def\sS{{\mathbb{S}}}

\DeclareMathOperator*{\argmax}{arg\,max}

\usepackage{enumitem}
\usepackage{symbols}

\definecolor{cvprblue}{rgb}{0.21,0.49,0.74}
\definecolor{lightcarminepink}{rgb}{0.9, 0.4, 0.38}
\usepackage[pagebackref,breaklinks,colorlinks,citecolor=Black,linkcolor=lightcarminepink]{hyperref}

\usepackage{tabularx}
\usepackage{algorithm}
\usepackage{algorithmic}
\usepackage{tikz}
\usetikzlibrary{matrix}

\usepackage{caption}
\usepackage{amsmath}
\usepackage{amssymb}
\definecolor{maroon}{cmyk}{0,0.87,0.68,0.32}
\usepackage{colortbl}

\usepackage[utf8]{inputenc} %
\usepackage[T1]{fontenc}    %
\usepackage{url}            %
\usepackage{booktabs}       %
\usepackage{amsfonts}       %
\usepackage{nicefrac}       %
\usepackage{microtype}      %
\usepackage{xcolor}         %
\usepackage{bbm}

\usepackage{graphicx}
\usepackage{booktabs}
\usepackage{multirow}       %
\usepackage{bbm}
\usepackage{subcaption}

\usepackage[utf8]{inputenc} %
\usepackage[T1]{fontenc}    %
\usepackage{hyperref}       %
\usepackage{url}            %
\usepackage{booktabs}       %
\usepackage{amsfonts}       %
\usepackage{nicefrac}       %
\usepackage{microtype}      %
\usepackage{xcolor}         %

\def\BibTeX{{\rm B\kern-.05em{\sc i\kern-.025em b}\kern-.08em
    T\kern-.1667em\lower.7ex\hbox{E}\kern-.125emX}}
\begin{document}

\title{Semantically Consistent Language Gaussian Splatting\\ for 3D Point-Level Open-vocabulary Querying}

\author{Hairong Yin$^{1}$, Huangying Zhan$^{2}$, Yi Xu$^{2}$, Raymond A. Yeh$^{1}$
\thanks{$^{1}$Department of Computer Science, Purdue University, USA.
        {\tt\small \{yin178, rayyeh\}@purdue.edu}.}%
\thanks{$^{2}$Goertek Alpha Labs, USA. {\tt\small \{huangying.zhan, yi.xu\}@goertekusa.com}.}
\thanks{Paper website: $\href{https://evelinyin.github.io/seconGS/}{https://evelinyin.github.io/seconGS/}$}
}

\maketitle

\begin{abstract}
Open-vocabulary 3D scene understanding is crucial for robotics applications, such as natural language-driven manipulation, human-robot interaction, and autonomous navigation. Existing methods for querying 3D Gaussian Splatting often struggle with inconsistent 2D mask supervision and lack a robust 3D point-level retrieval mechanism. In this work, (i) we present a novel point-level querying framework that performs tracking on segmentation masks to establish a semantically consistent ground-truth for distilling the language Gaussians; (ii) we introduce a GT-anchored querying approach that first retrieves the distilled ground-truth and subsequently uses the ground-truth to query the individual Gaussians. 
Extensive experiments on three benchmark datasets demonstrate that the proposed method outperforms state-of-the-art performance. Our method achieves an mIoU improvement of +4.14, +20.42, and +1.7 on the LERF, 3D-OVS, and Replica datasets. These results validate our framework as a promising step toward open-vocabulary understanding in real-world robotic systems.
\end{abstract}
 
\section{Introduction}
\label{sec:intro}
Querying open-vocabulary objects in 3D scenes, \ie, identifying and isolating scene components based on natural language descriptions, is a fundamental capability necessary to advance robotic perception and interaction. To evaluate open-vocabulary querying, recent works~\cite{feature3dgs, langsplat, opengaussian, lerf, supergseg, ye2025gaussian, ji2025fastlgs} have formulated the task in two steps: {\bf (a)} creating a 3D scene representation augmented with language-aligned features, and {\bf (b)} querying this representation effectively using language. Existing works address {\bf (a)} by distilling language embeddings from foundation 2D vision-language models (VLMs)~\cite{lseg, clip} into point clouds~\cite{conceptfusion, xu2024embodiedsam, mohiuddin2024opensu3d}, NeRF~\cite{nerf,martin2021nerf,xie2022neural} and 3D Gaussians~\cite{3dgs,yu2024mip,chen2024survey}. To address {\bf (b)}, they perform retrieval by thresholding on the cosine similarity between the text query embedding and the distilled language embedding. 

While many methods have explored this task, a distinction lies in their output querying format. Approaches,~\eg, LangSplat~\cite{langsplat}, that produce \textit{2D segmentation masks} are insufficient for robotics, as downstream tasks like motion planning, grasp synthesis, and collision avoidance often operate directly on a 3D representation. This led to the development of \textbf{point-level querying}~\cite{ye2025gaussian, opengaussian, supergseg}, which directly retrieves a subset of the underlying \textit{3D primitives}. In the case of 3D Gaussian Splatting~\cite{3dgs}, this involves retrieving the relevant subset of Gaussians from the scene. The resulting 3D selection provides a direct and actionable representation for a robot's planning and control. %

\begin{figure}[t]
\input{figs/teaser}
\vspace{-0.1cm}
\end{figure}
In this work, we identify two main shortcomings of LangSplat~\cite{langsplat} (initially proposed for 2D querying) %
when applied to 3D point-level querying. 
First, we observe inconsistency in the distillation ground-truth language embeddings constructed by LangSplat. That is, the embeddings are different for the same object instance across different frames; see in~\figref{fig:teaser}. To address this, we propose a tracking-based distillation process and aggregate the language embedding into a consistent ground-truth. 

\begin{figure}[t]
\centering
\includegraphics[width=0.6\linewidth]{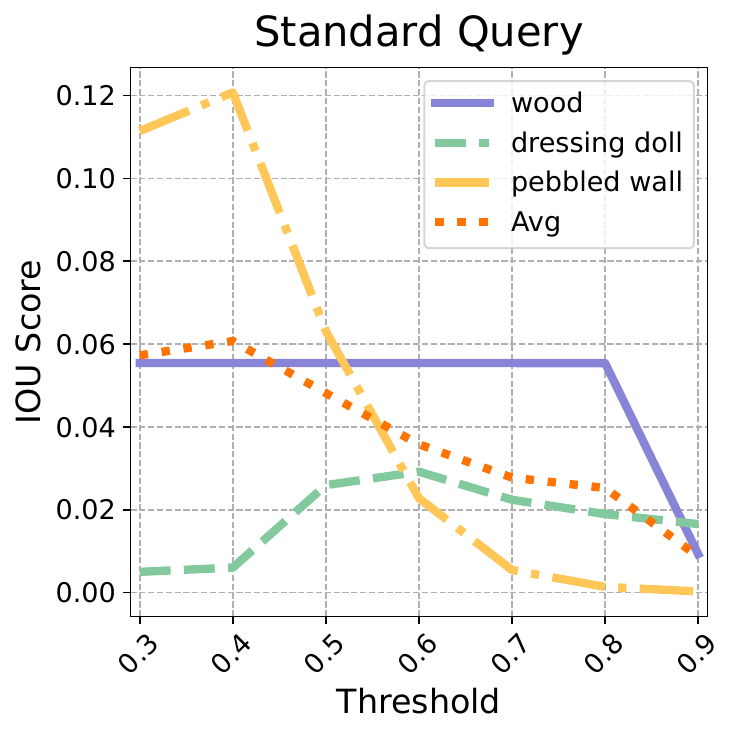}
\vspace{-0.2cm}
\caption{IoU metric per query vs. cosine similarity thresholds for the standard querying method. We observe that it does not have a consistent optimal threshold for all queries.}
\label{fig:standard_query}
\vspace{-0.4cm}
\end{figure}

Next, the second challenge lies in the querying phase. LangSplats' querying approach thresholds the similarity between query text vectors and learned point-wise language embeddings; however, choosing an appropriate threshold across different text queries is challenging. As shown in~\figref{fig:standard_query}, the optimal thresholds are \textit{not the same} across all objects. To mitigate this difficulty, we propose a novel Ground-Truth Anchored (GT-Anchored) querying method, which computes the threshold relative to, ``anchored'', ground-truth (GT) used in the distillation process instead of directly with the text query.

Empirically, we conduct experiments over three datasets: LERF~\cite{lerf}, 3D-OVS~\cite{3dovs}, and Replica~\cite{replica}, demonstrating that our method outperforms the state-of-the-art method in terms of mIoU by $+4.14, +20.42$, and $+1.74$, respectively. A detailed ablation study is conducted to verify the effectiveness of the proposed components. 

{\noindent\bf Our contributions are as follows:}
\begin{itemize}[topsep=0pt, leftmargin=16pt]
    \setlength{\itemsep}{0.0pt}
    \setlength{\parskip}{2.5pt}
    \item We introduce tracking for generating semantic and 3D-consistent ground-truth to train language-aware Gaussians, which improves the distillation quality.
    \item With this improved 3D language Gaussians, we propose an effective GT-anchored querying process by leveraging the created consistent ground-truth to alleviate the aforementioned challenge of selecting a suitable threshold.
    \item Extensive experiments across three datasets demonstrate that our approach outperforms existing open-vocabulary 3D querying methods. %
\end{itemize}

\section{Related Work}
\label{sec:rel_work}

\subsection{Vision foundation models.}
Several open-sourced foundation models~\cite{clip, dino, sam1, sam2} have become the bedrock of many works~\cite{conceptfusion, cen2023segment, ji2025fastlgs}. These foundation models' capabilities can either be used directly or their features can be distilled into another model. 
In language and vision, CLIP~\cite{clip} is a model that is capable of encoding images and natural language text to the same embedding space. 
Using this joint embedding space, they demonstrate zero-shot capability for image classification, which is later generalized to segmentation~\cite{zhou2022extract} and language segmentation (LSeg)~\cite{lseg}. 

In the area of image segmentation, the Segment Anything Model (SAM)~\cite{sam1} is a notable foundation model. 
SAM's segmentation capabilities have been adapted and extended to 3D tasks. 
Recent works~\cite{cen2023segment, kim2024garfield} have leveraged SAM to integrate semantic information into NeRFs, enabling the extraction of 3D masks for target objects. 
Other approaches~\cite{liu2024segment,he2025segpoint} have incorporated semantic features into point clouds, enhancing object representation and segmentation in 3D. 
The process in 3D Gaussian Splatting~\cite{3dgs} has further motivated studies~\cite{ye2025gaussian, ji2024segment} that focus on object representation in both 3D and 4D, including advancements in interactive segmentation and object tracking. 
Recently, SAM2~\cite{sam2} extended SAM's capability to tracking of masklets, \ie, consistent masks across both space and time. %

\subsection{Open-vocabulary 3D scene understanding.}
With advancements in 3D scene representation, there is a surge in interest in incorporating semantics/language into 3D representation.
LERF~\cite{lerf} and other works~\cite{3dovs,shen2023distilled,ye2023featurenerf} distilled features from DINO~\cite{dino} and CLIP~\cite{clip} to learn a NeRF~\cite{nerf}, or leveraged 2D annotations~\cite{zhi2021place,siddiqui2023panoptic} to construct feature/language fields. Others~\cite{conceptfusion, xu2024embodiedsam, mohiuddin2024opensu3d, yamazaki2024openfusion} distill knowledge from these language-rich models into point clouds or voxels, enabling open-vocabulary 3D scene understanding.
In more recent works~\cite{langsplat,feature3dgs,ye2025gaussian,opengaussian,semanticgs}, there is a shift towards 3D Gaussian Splatting~\cite{3dgs}. 

More closely related to our work is LangSplat~\cite{langsplat}, which augments 3D scenes with language features distilled from CLIP, enabling natural language querying on the renderings of the 3D scene. Gaussian Grouping~\cite{ye2025gaussian} jointly performs 3D reconstruction and segmentation of open-world objects. It generates 2D masks using SAM and associates them across frames through a zero-shot tracker. The method also incorporates a custom loss that enforces 3D consistency. However, the method leverages the tracking information differently from ours by learning a group ID for each Gaussian point, which relies on a complicated grouping loss. Also, it selects target objects using only the semantics of the first frame, which could lead to potential query failures. OpenGaussian~\cite{opengaussian} introduces new loss functions that leverage inter- and intra-mask smoothness relationships, along with a codebook-based clustering method to improve instance-level association of 3D points. 

Differently, our approach does not rely on new loss functions. Instead, our tracking-based method extracts masklets to construct consistent ground-truth supervision and introduces a novel GT-anchored querying procedure.

\begin{figure*}[t]
\centering
\includegraphics[width=0.9\linewidth]{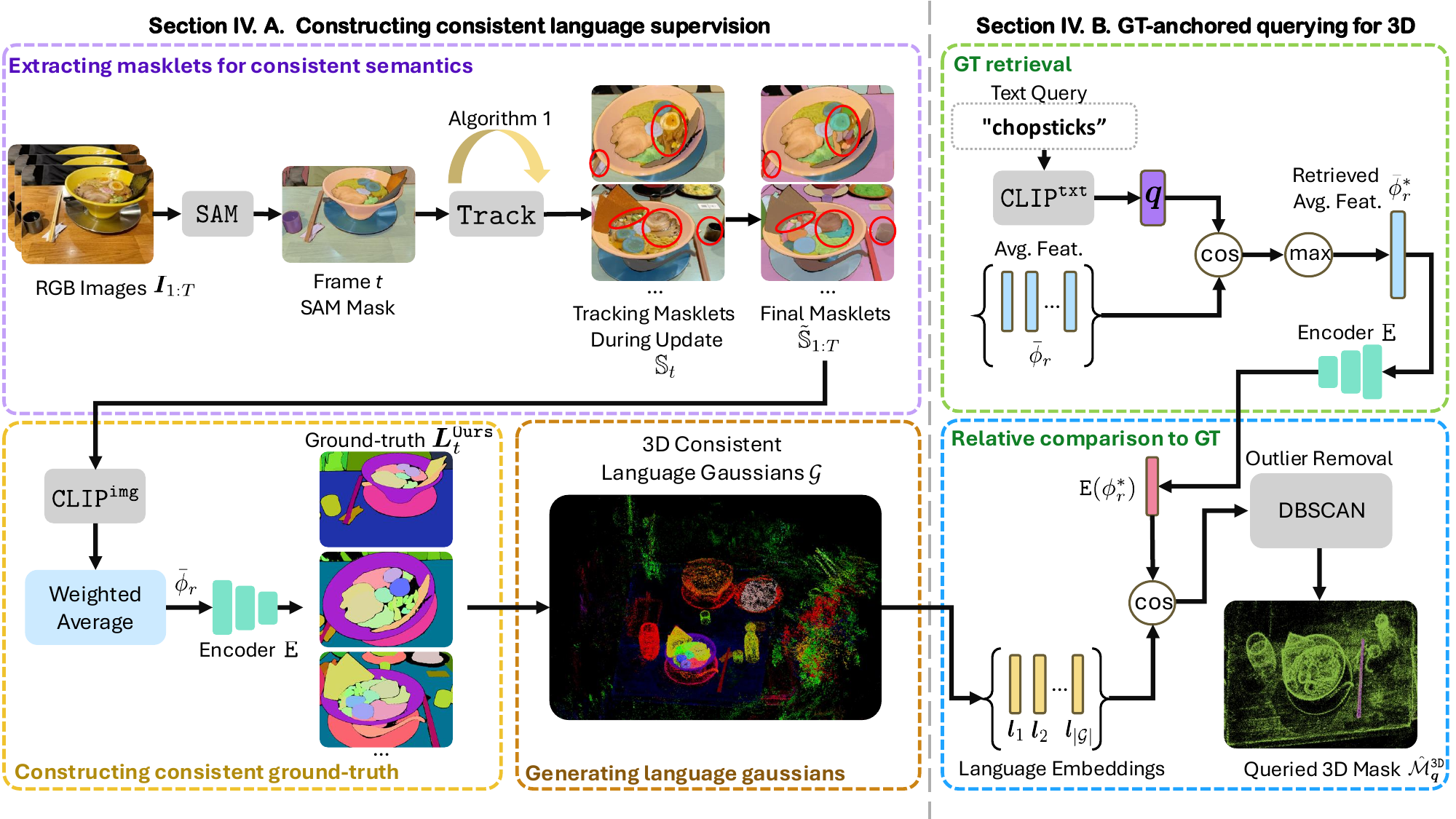}
\vspace{-.22cm}
\caption{Overview of the proposed method. In~\secref{sec:consistent}, 
we present a masklet extraction algorithm (\algref{alg:method}) that leverages Segment Anything Models to generate consistent ground-truth $\mL_t^{\tt Ours}$ for training the language parameters ${\vl_1, \dots, \vl_{|\gG|}}$.
In~\secref{sec:query}, we discuss the GT-anchored retrieval procedure. 
Rather than directly querying the language parameters $\vl_i$ with the query vector $\vq$, we first retrieve the features $\bar{\phi}_r$ that are used to construct the ground truth $\mL_t^{\tt Ours}$. 
Then we query the 3D language Gaussian using the encoded feature $\texttt{E}(\bar{\phi}^*_r)$, followed by an outlier removal using DBSCAN~\cite{dbscan} to obtain the final result. 
}
\vspace{-.3cm}
\label{fig:pipeline}
\end{figure*}

\section{Preliminaries}\label{sec:prelim}
We review LangSplat and introduce the necessary notation.

{\noindent\bf LangSplat}~\cite{langsplat} 
represents a 3D scene with a set of 3D Gaussians $\gG = \{\vg_i\}$, where each Gaussian $\vg_i$ is associated with the parameters 
\bea\label{eq:gauss}
\vg_i \triangleq (\bm\mu_i, \bm\Sigma_i, \vc_i, \alpha_i, \vl_i)
\eea
corresponding to the 3D location, covariance matrix, color, opacity, and a language embedding. Different from a regular 3D Gaussian splatting~\cite{3dgs}, each of the Gaussians (\equref{eq:gauss}) includes a language embedding $\vl_i \in \sR^D$ to encode the semantics of a 3D scene. 

This language embedding can then be rendered into a language field $\hat\mL_\pi \in \sR^{H \times W \times D}$, where $H$ and $W$ correspond to the height and width of the rendered image at a camera pose $\pi$. This is done by using a tile-based rasterization~\cite{zwicker2001ewa}, just as one would for colors. 
The language feature at each pixel $\vp$ is computed as
\bea
\hat\mL_\pi[\vp] = \sum_{i \in \gT} \vl_i f_i^{\tt 2D} (\vp)\prod_{j=1}^{i-1} (1-f_j^{\tt 2D}(\vp)),
\eea 
where $f_i^{\tt 2D}$ represents the the projected 2D contribution of a 3D Gaussian $\vg_i$, and $\gT$ is the set of Gaussians in a tile. 

{\noindent\bf LangSplat training.} 
Let $\hat{\mL}_{\pi_t}$ denote the rendering of the scene from the camera pose $\pi_t$ associated with image $\mI_t$.
LangSplat trains language embedding $\vl_i$ by minimizing the L1 loss between the rendering of the language feature $\hat\mL_\pi$ and a ``ground-truth'' language feature $\mL_t$, \ie,
\bea\label{eq:lang_train}
\min_{\{\vl_i \forall i\}} \sE_{\mI_t} \left[ \text{L1}(\hat\mL_{\pi_t},\mL_t) \right].
\eea
Importantly, how one designs this ``ground truth'' significantly affects the query performance.

In LangSplat, the ground-truth is distilled from a pretrained CLIP~\cite{cherti2023reproducible} with the help of the segment anything model (SAM)~\cite{sam1}. Given an RGB image $\mI$, SAM extracts a set of non-overlapping segmentation masks 
$
\sS_{\mI} \triangleq \{\gS_{r}\} = \texttt{SAM}(\mI),
$
where each $\gS_{r}$ corresponds to the segmented mask of a region $r$. Here, we broadly use the term ``region'' to mean a set of related pixels, \eg, an object or a subpart of an object. 

Next, a CLIP embedding $\phi_{r,t}$ is extracted for each region at time $t$, 
LangSplat~\cite{langsplat}
masks the image, then passes it to CLIP's image encoder $\texttt{CLIP}^{\text{img}}: \sR^{H\times W \times 3} \rightarrow \sR^D$, \ie, 
\bea
\phi_{r,t} \triangleq \texttt{CLIP}^{\text{img}}(\mI_t \odot \mS_r),
\eea
where $\mS_r$ corresponds to the mask $\gS_r$ in matrix form, and $\odot$ denotes an element-wise multiplication. 
These extracted region features are then placed back to %
their respective regions to form the ground-truth
\bea \label{eq:langsplat_gt}
 \mL_t[(h,w)] \triangleq \phi_{r,t} \text{ if }(h,w) \in \gS_r ~\forall r.
\eea
As the language embedding $\mL$ is distilled from a pretrained CLIP embedding, the query vector $\vq$ is extracted from the pretrained CLIP text encoder. This ensures that both the query vector and the 3D language embeddings are in the same space, allowing for retrieval.

Lastly, we note an implementation detail, 
LangSplat~\cite{langsplat} trains an autoencoder, consisting of an encoder $\texttt{E}$ and a decoder $\texttt{D}$, to reduce the dimensions of the CLIP features, where $\texttt{E} \circ \texttt{D}$ approximates an identity function. With this autoencoder, all the aforementioned formulations can be done in a lower-dimensional space to save GPU memory.
However, this dimensionality reduction introduces a trade-off: language features for the same object become less consistent across views due to compression, as illustrated in Fig.~\ref{fig:teaser}.

{\noindent\bf Open-vocabulary 3D (point-level) querying.} %
OpenGaussian~\cite{opengaussian} proposes to
directly query the 3D Gaussians with natural language. 
Formally, given a trained 3D scene $\gG$ and a query $\vq$, the task is to predict 
a \textit{``3D mask''}
\bea
\gM^{\tt 3D}_\vq = \{\vg_i\} \subseteq \gG,
\eea
that indicates whether each Gaussian $\vg_i$ is relevant to a text query $\vq$. In other words, a \textit{point-level} querying on the 3D scene's representation rather than on a \textit{rendered image} of a 3D scene.

{\noindent\bf Tracking framework.} %
Given a sequence of frames $\mI_{t_1:t_2}$, and a set of candidates $\tilde\gB_{t_1}$ indicating the regions in frame $t_1$ that we wish to track and segment, a tracking model extracts a set $\tilde{\sS}_{\mI_{t_1:t_2}}$ of non-overlapping masks across both \textit{space and time}, \ie, ``masklets''
\bea
\tilde{\sS}_{\mI_{t_1:t_2}} \triangleq \{\tilde\gS_{r} \} = \texttt{Track}(\mI_{t_{1}:t_{2}}, \gB_{t_1}),
\eea
where each $\tilde\gS_{r}$ is a set containing voxels $(t,h,w)$ that are associated with the region $r$. Following the same syntax for a 2D binary mask, we use the tensor $\tilde\mS_r\in \{0,1\}^{T\times H\times W}$ to represent the masklet $\tilde\gS_{r}$ in set notation.

\section{Method}
\label{sec:method}

Following LangSplat, we use a set of 3D Gaussians augmented with language embeddings to represent a scene. Differently, we propose a novel method for constructing ground-truths that are more semantically consistent and robust across various 3D viewpoints (\secref{sec:consistent}). This approach 
helps to train better language embeddings for querying. We then introduce a querying method tailored for our learned embeddings %
(\secref{sec:query}). See~\figref{fig:pipeline} for a visual overview.

\begin{figure}[t]
    \centering
    \setlength{\tabcolsep}{2pt}
    \begin{tabular}{cc}
        \includegraphics[width=.43\linewidth]{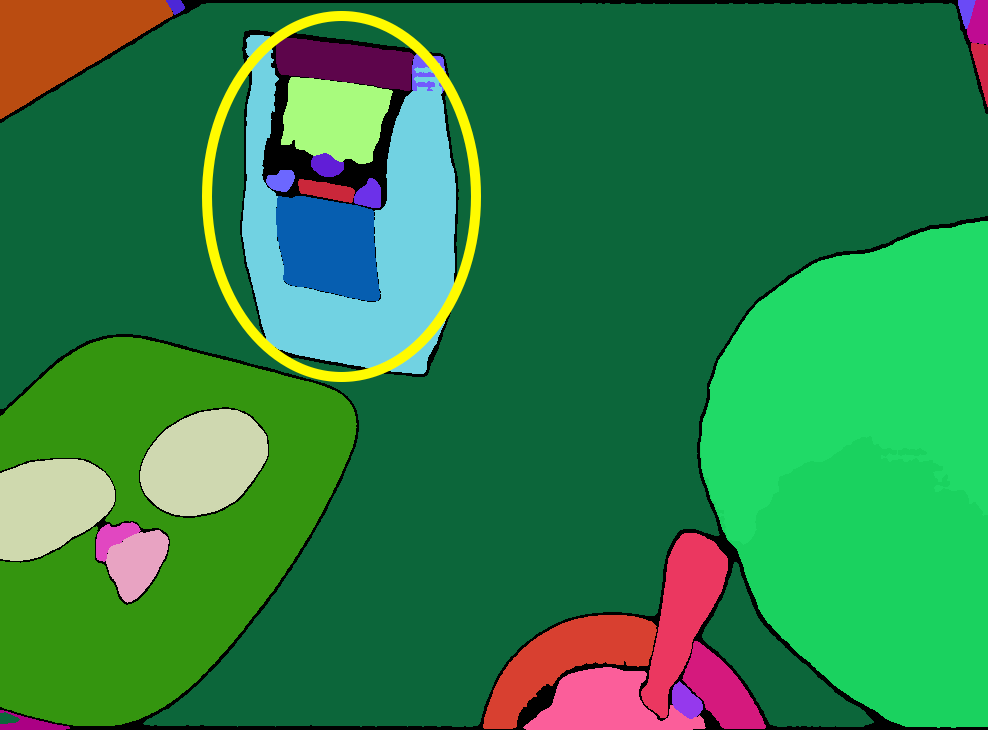} & 
        \includegraphics[width=.43\linewidth]{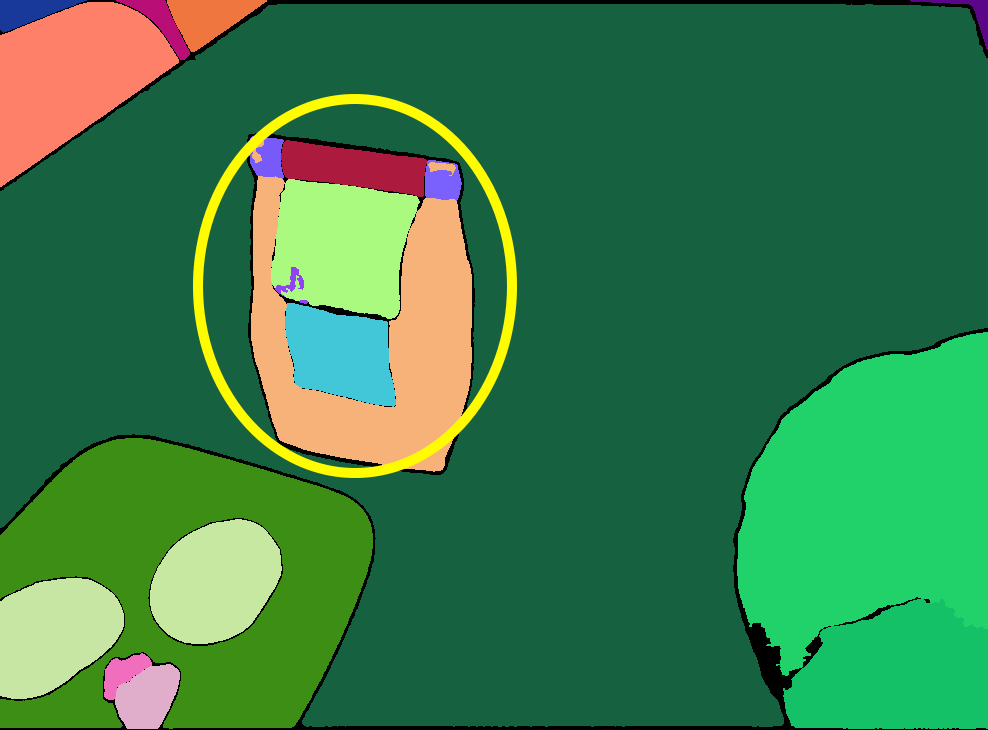} \\
        \includegraphics[width=.43\linewidth]{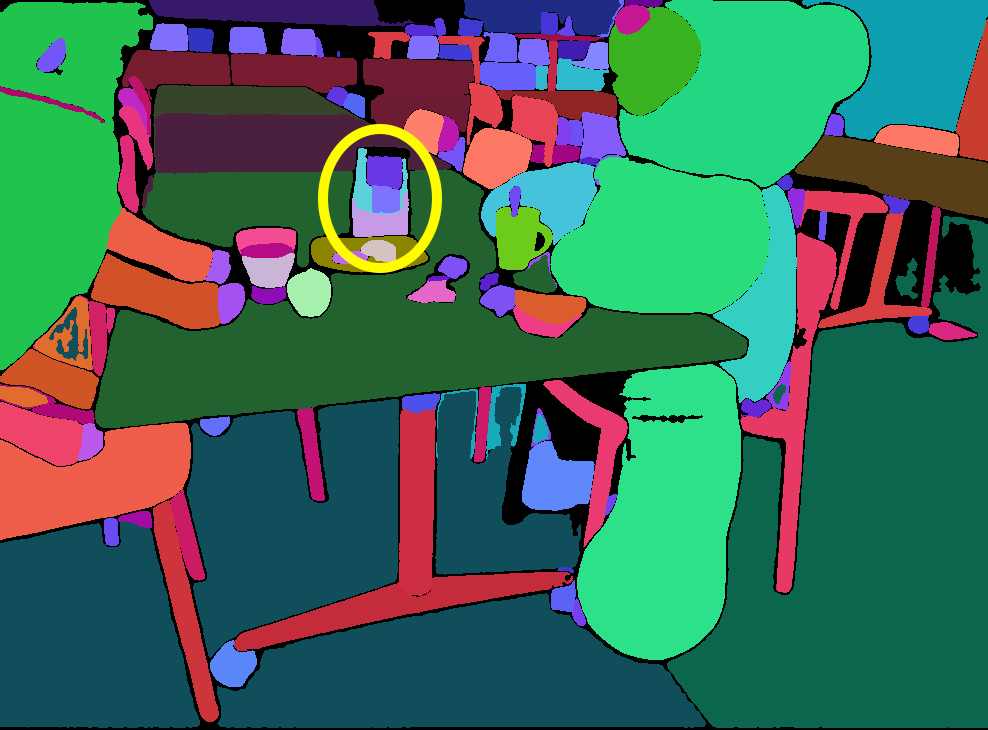} & 
        \includegraphics[width=.43\linewidth]{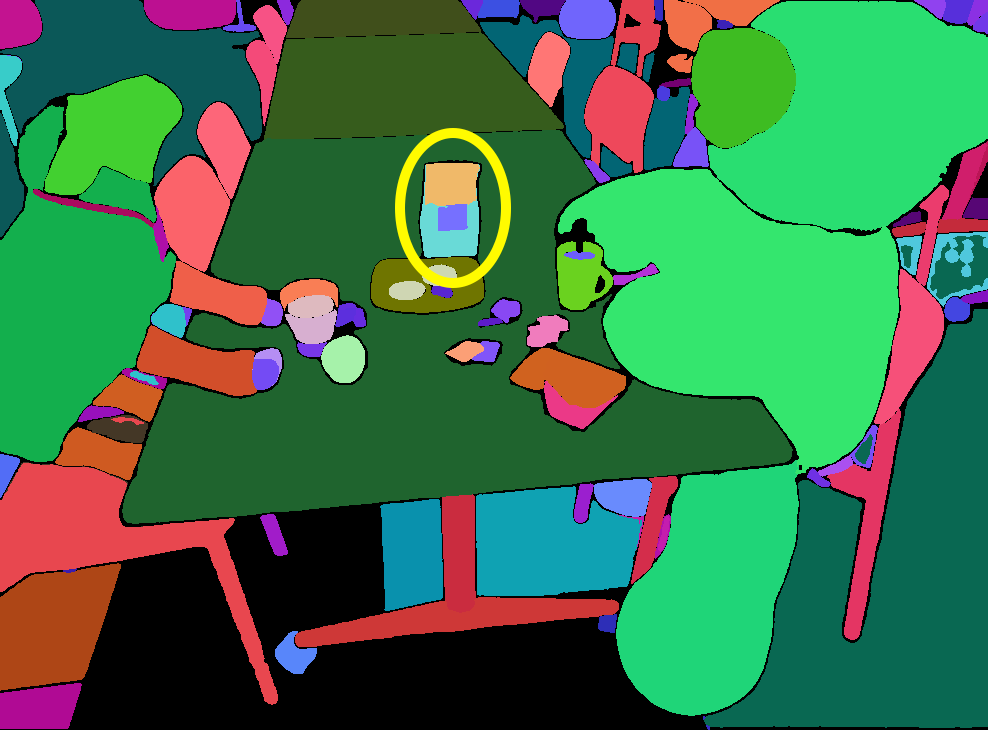}
        \end{tabular}
    \vspace{-0.2cm}
    \caption{Visualization of the ground-truth $\mL_t$ constructed by LangSplat~\cite{langsplat}. We observed that the semantics are not consistent across viewpoints,~\eg, the circled bag of coffee.}
    \label{fig:lang_incon}
    \vspace{-0.35cm}
\end{figure}

\subsection{Constructing consistent language supervision} 
\label{sec:consistent}
 Given a sequence of frames $[\mI_1, \dots \mI_T]$ with camera poses, we aim to construct a better ground-truth feature $\mL^{\tt Ours}_t$ for each of the frames to train LangSplat's parameters by minimizing the objective function in~\equref{eq:lang_train}. This improved ground-truth is obtained by ensuring that all pixels within the same region, as identified by the chosen tracking model (SAM2~\cite{sam2}), share the same CLIP embedding. In other words, the supervision will be semantically consistent up to the quality of the extracted masklets from SAM2.

{\bf\noindent Inconsistent semantics from LangSplat.}
The main shortcoming of the ground-truth feature $\mL^{\tt}_t$ created by LangSplat is its potential inconsistency across different views. In~\figref{fig:lang_incon}, we visualize these features constructed by LangSplat, where similar colors indicate higher feature similarity. 
As shown, the bag of cookies (circled in yellow) exhibits significant mask variations across the views, indicating inconsistent supervision. 
This inconsistency arises because SAM segments each image independently, potentially selecting different regions. Hence, the ground-truth features derived from these segmentations are also inconsistent.

To address this inconsistency, we construct the ground truth $\mL^{\tt Ours}_t$ by additionally leveraging the tracking capabilities of SAM2. Importantly, we aim to design $\mL^{\tt Ours}_t \in \sR^{H\times W \times D}$ such that each pixel's feature remains consistent across frames when it belongs to the same region extracted by SAM2.

\begin{figure}[t]
\small
\centering
\begin{minipage}{\linewidth}
\vspace{-.3cm}
\begin{algorithm}[H]
   \caption{Extracting regions with SAM and Tracking}
   \label{alg:method}
    \begin{algorithmic}[1]
       \STATE {\bfseries Input:} Image sequence $\mI_{1:T}$, segmentors $\texttt{SAM}$ and Tracker $\texttt{Track}$, threshold $\kappa$
       \STATE $\tilde\sS_{1:T}\leftarrow\{\}$ \tt {\color{gray}\# Tracked masklets}
       \FOR{$t \in \{1, \dots, T\}$}
        \STATE $\sS_{\mI_t} = \texttt{SAM}(\mI_t)$
        \STATE {\color{gray} \tt{\# Check if tracked.}}
        \FOR{$\mS_r \in \sS_{\mI_t}$}
        \FOR{$\tilde\mS_{r'} \in \tilde\sS_{1:T}$}
        \IF{$\text{IoU}(\tilde\mS_{r'}[t-1],\mS_r) > \kappa$}
        \STATE $\sS_{\mI_t} \leftarrow \sS_{\mI_t} \backslash \{\mS_r\}$
        \ENDIF
        \ENDFOR
        \ENDFOR
        \STATE {\color{gray} \tt{\# Adding untracked masklets}}
        \STATE $\gB \leftarrow \texttt{RegionFromMask}(\sS_{\mI_t})$
        \STATE $ \tilde\sS_{1:T} \leftarrow \tilde\sS_{1:T} \cup \texttt{Track}(\mI_{1:T}, \gB)$
       \ENDFOR
       \STATE {\bfseries Output:} $\tilde\sS_{1:T}$
    \end{algorithmic}
    \end{algorithm}
\end{minipage}
\vspace{-0.3cm}
\end{figure}

{\noindent\bf Extracting masklets for consistent semantics.} Reviewed in~\secref{sec:prelim}, a tracking module takes a sequence of images and regions of interest as input to track masks of the same region. In LangSplat, SAM is used to extract the initial candidates, \ie, for each frame $\mI_t$, SAM proposes a set of regions $\sS_{\mI_t}$. Starting from the first frame, we check if a proposed region has already been tracked by comparing the mIoU of the SAM mask with the tracked masks and applying a threshold. If the proposed region has not been tracked, we run the tracking model and add the output masklets to the set of tracked masklets $\tilde\sS_{1:T}$. These steps are summarized in~\algref{alg:method}.

{\noindent\bf Constructing consistent ground-truth.} 
With the set of masklets $\tilde\sS_{1:T}$ extracted, we create a consistent ground-truth by aggregating the CLIP embedding $\bar\phi_r$ for each masklet $\tilde\gS_r \in \tilde \sS_{1:T}$. This is done by masking out the image $\mI_t$ using the extracted masklet and then passing it to CLIP's image encoder:
\bea
\bar\phi_r = \sum_{t=1}^{T} \omega_t \cdot \texttt{CLIP}^{\tt img}(\mI_t \odot \tilde\mS_r[t]),
\eea
where $\tilde\mS_r$ denotes the masklet $\tilde\gS_r$ represented in a tensor, and $\omega_t$ denotes the ratio of pixels in $\tilde\mS_r[t]$ to the total pixel count in $\tilde\mS_r$.  In other words, the embedding from each view is weighted proportionally to the number of pixels in the segmentation. 

The main intuition behind this design is that averaging reduces variance. The proposed $\bar\phi_r$ is more consistent than the individual $\phi_{r,t}$ used in LangSplat when used as supervision for the distillation. Furthermore, the weighting scheme helps to suppress the contribution of small regions that often contain noisier language embeddings, \ie, we consider the reliability of individual features.

\begin{table*}[t]
    \small
    \setlength{\tabcolsep}{2pt}
    \centering
    \caption{Quantitative results on LERF. %
    For OpenGaussian, we report the numbers from their paper.
    }
    \vspace{-0.15cm}
        \resizebox{\linewidth}{!}{%
    \begin{tabular}{l | c c c c c | c c c c c | c c c c c}
     \specialrule{.10em}{.01em}{.01em}
     \multirow{2}{*}{Methods} &
     \multicolumn{5}{c|}{mIoU$\uparrow$} &
     \multicolumn{5}{c|}{mAcc$\uparrow$} & 
     \multicolumn{5}{c}{Loc. Acc$\uparrow$} \\
     & figurines & ramen & teatime  & kitchen & \textbf{Avg} & figurines & ramen & teatime  & kitchen & \textbf{Avg} & figurines & ramen & teatime  & kitchen & \textbf{Avg} \\
    \hline
        LangSplat-m~\cite{langsplat}  & 12.43 & 6.39 & 20.60 & 17.58 & 14.25 & 21.43 & 7.04 & 37.29 & 18.18 & 20.99  & 5.36 & 0.00 & 3.39 & 4.55 & 3.33 \\
         GSGroup.-m~\cite{ye2025gaussian} & 7.75 & 8.80 & 10.94 & 16.29 & 10.95 & 10.71 & 9.86 & 10.17 & 27.27 & 14.50 & 10.71 & 2.82 & 5.08 & 4.55 & 5.79 \\
         OpenGauss.~\cite{opengaussian}  & 39.29 & 31.01 & \textbf{60.44} & 22.7 & 38.36 & 55.36 & 42.25 & \textbf{76.27} & 31.82 & 51.43 & - & - & - & - & -\\
         \rowcolor{gray!10}
        \textbf{Ours} & \textbf{58.91} & \textbf{37.85} & 43.57 & \textbf{29.67} & \textbf{42.50}  & \textbf{82.14} & \textbf{61.97} & 54.24 & \textbf{50.00} & \textbf{62.09} & \textbf{82.14} & \textbf{61.97} & \textbf{62.71} & \textbf{40.91} & \textbf{61.93}\\
    \specialrule{.10em}{.01em}{.01em}
    \end{tabular}
    }
    \vspace{-0.1cm}
    \label{tab:lerf}
\end{table*}

\begin{table*}[t]
    \small
    \setlength{\tabcolsep}{2pt}
    \centering
    \caption{Quantitative results on the 3D-OVS dataset.}
    \vspace{-0.2cm}
    \resizebox{\linewidth}{!}{%
    \begin{tabular}{l | c c c c c c | c c c c c c | c c c c c c}

     \specialrule{.10em}{.01em}{.01em}
     \multirow{2}{*}{Methods} &
     \multicolumn{6}{c|}{mIoU$\uparrow$} & 
     \multicolumn{6}{c|}{mAcc$\uparrow$} & 
     \multicolumn{6}{c}{Loc. Acc$\uparrow$} \\
     & bed & bench & lawn & room & sofa  & \textbf{Avg} & bed & bench & lawn & room & sofa  & \textbf{Avg} & bed & bench & lawn & room & sofa  & \textbf{Avg} \\
    \hline
    
        LangSplat-m~\cite{langsplat}  & 29.83 & 17.38 & 33.64 & 23.35 & 24.64 & 25.77  & 43.33 & 28.57 & 63.33 & 33.33 & 43.33 & 42.38 & 3.00 & 48.57 & 40.00 & 43.33 & 43.33 & 35.64 \\
        GSGroup.-m~\cite{ye2025gaussian} & 48.51 & 35.49 & 65.13 & 46.39 & 29.86 & 45.08 & \textbf{100.00} & 71.43 & \textbf{100.00} & 76.67 & 40.00 & 77.62 & \textbf{96.67} & 42.86 & \textbf{100.00} & 53.33 & 43.33 & 67.24 \\
        OpenGauss.~\cite{opengaussian} & 48.50 & 46.02 & 64.63 & 47.60 & 44.06 & 50.16  &\textbf{100.00} & 57.14 &\textbf{100.00}  & \textbf{83.33} & \textbf{66.67} & 81.43 & 23.33 & 37.14 & 20.00 & 50.00 & 56.67 & 37.43\\
        \rowcolor{gray!10}
        \textbf{Ours} & \textbf{56.81} & \textbf{87.58} & \textbf{87.12} & \textbf{64.70} & \textbf{56.70} & \textbf{70.58} & 66.67 & \textbf{100.00} & \textbf{100.00} & \textbf{83.33} & \textbf{66.67} & \textbf{83.33} &  83.33 & \textbf{100.00} & \textbf{100.00} & \textbf{83.33} & \textbf{100.00} & \textbf{93.33}\\
    \specialrule{.10em}{.01em}{.01em}
    \end{tabular}
    }
    \label{tab:3dovs}
    \vspace{-0.1cm}
\end{table*}

\begin{table*}[t]
    \small
    \setlength{\tabcolsep}{2pt}
    \centering
    \caption{Quantitative results on Replica dataset. To compute mIoU and mAcc using the ground-truth point clouds, we skip the densification stage when training 3D Gaussian Splatting for all methods.
    }
        \resizebox{\linewidth}{!}{%
    \begin{tabular}{l | c c c c c c c c c | c c c c c c c c c }
     \specialrule{.10em}{.01em}{.01em}
     \multirow{2}{*}{Methods} &
     \multicolumn{9}{c|}{mIoU$\uparrow$} &
     \multicolumn{9}{c}{mAcc$\uparrow$} \\
     & office0 & office1 & office2  & office3 & office4 & room0 & room1 & room2  & \textbf{Avg} & office0 & office1 & office2  & office3 & office4 & room0 & room1 & room2 & \textbf{Avg} \\
    \hline
        LangSplat-m~\cite{langsplat}& 2.43 & 2.1 & 5.68 & 4.65 & 1.49 & 3.86 & 4.08 & 0.92 & 3.15 & 11.09 & 1.36 & 10.7 & 13.99 & 2.37 & 12.82 & 12.24 & 10.05 & 9.33 \\
        GSGroup.-m~\cite{ye2025gaussian} & 19.58 & 0.00 & 32.77 & 10.18 & 30.29 & 13.08 & 17.81 & 17.06 & 17.60 & 38.42 & 0.00 & \textbf{74.48} & 26.17 & 45.67 & 36.21 & 31.57 & 24.17 & 34.59 \\
        OpenGauss.~\cite{opengaussian} & 17.20 & \textbf{23.13}  & \textbf{43.72}  & \textbf{42.36} & 61.33 & 31.45 & 40.36 & 42.14 & 37.71
        &36.54 & 35.11 & 66.38 &42.64 & 69.62 & 41.74 & 31.72 & 54.01 & 47.22\\ 
        \rowcolor{gray!10}
        \textbf{Ours} & \textbf{25.77}& 20.15 & 15.06 & 37.29 & \textbf{64.83} & \textbf{40.33} & \textbf{64.39} & \textbf{47.81} & \textbf{39.45}
        & \textbf{50.76} & \textbf{35.97} & 29.01 & \textbf{45.12} & \textbf{82.85} & \textbf{60.00} & \textbf{84.72} & \textbf{63.64} & \textbf{56.51}\\
    \specialrule{.10em}{.01em}{.01em}
    \end{tabular}
    }
    \vspace{-0.1cm}
    \label{tab:replica}
\end{table*}

To construct the ground truth $\mL^{\tt Ours}_t$ for a frame $\mI_t$, we place the averaged CLIP embedding into the pixel location of each frame according to the masklet. For all extracted masklets $\tilde\gS_r \in \tilde\gS_{1:T}$,
\bea
\mL^{\tt Ours}_t[(h,w)] = \bar\phi_r \text{ if } (t,h,w) \in \tilde\gS_r.
\eea
For pixels across views that are tracked and segmented into the same region, this construction of ground-truth assigns the same averaged CLIP embedding. Compared with LangSplat's ground-truth in~\equref{eq:langsplat_gt}, our construction is \textbf{shared across time} $t$, \ie, the language embedding $\vl_i$ in LangSplat receives consistent supervision across all relevant frames. %

\begin{figure*}
\centering
\setlength{\tabcolsep}{1pt}
\renewcommand{\arraystretch}{0.0}
\includegraphics[width=0.83\linewidth]{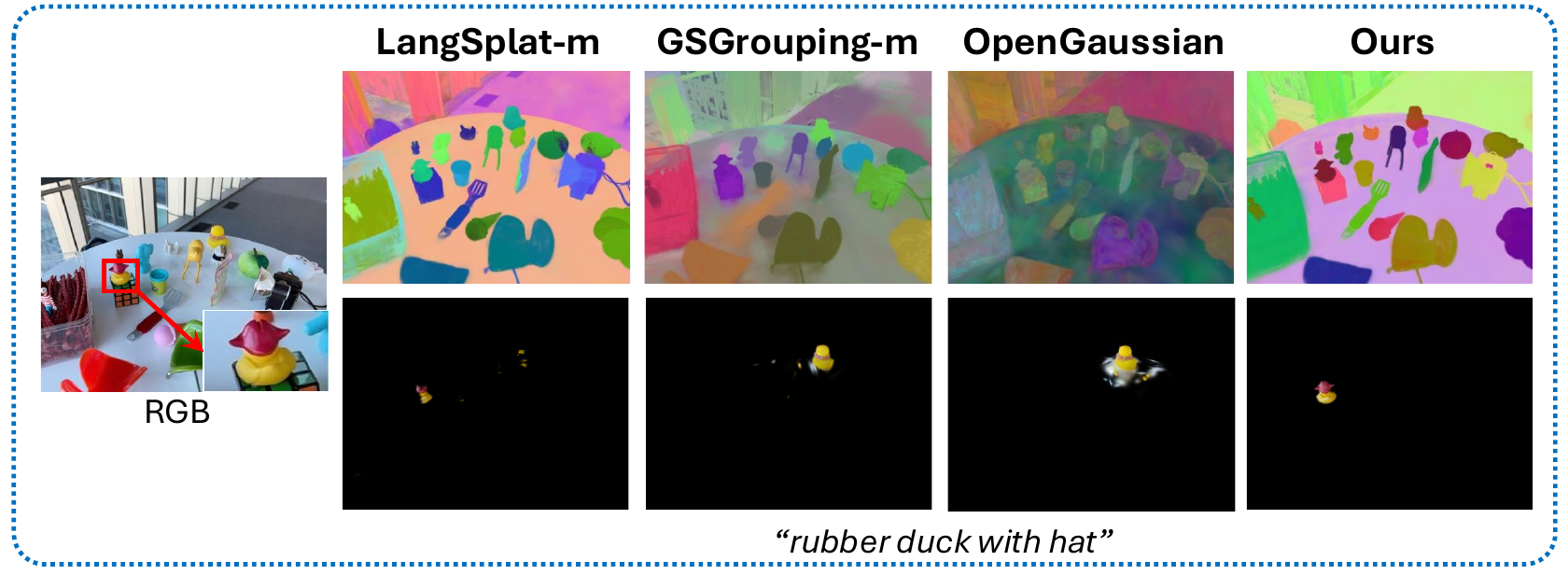}\\
\vspace{-0.2cm}
\caption{Qualitative results on LERF dataset of scene``ramen" and ``figurines". For each scene, the first row contains rendered language embeddings, and the second row contains 3D query results.
}
\vspace{-0.1cm}
\label{fig:lerf_qualitative}
\end{figure*}

\begin{figure*}[t]
\centering
\setlength{\tabcolsep}{1pt}
\renewcommand{\arraystretch}{0.0}
\includegraphics[width=0.83\linewidth]{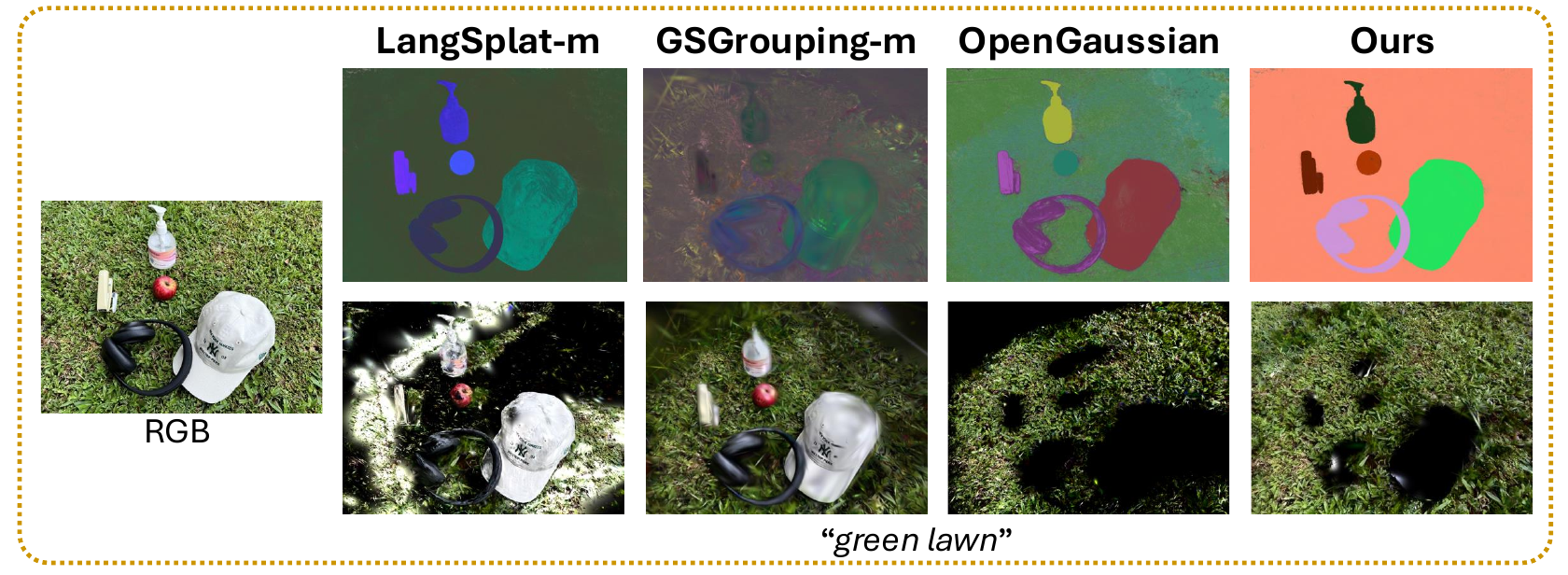}
\vspace{-0.2cm}
\caption{Qualitative results on 3D-OVS dataset for scene ``lawn". The first row contains rendered language embeddings, and the second row contains 3D query results for ``green lawn".
}
\vspace{-0.25cm}
\label{fig:3dovs_qualitative}
\end{figure*}

{\noindent\bf Autoencoder details.} As in LangSplat, to reduce the GPU memory usage,  we train a light-weight autoencoder consisting of an encoder $\texttt{E}$ and a decoder $\texttt{D}$. Differently, we encode the averaged CLIP embedding $\bar\phi_r$ into a low-dimensional latent space, \ie,%
$\texttt{E} ( \bar\phi_r  )$ is used as supervision to train the language embeddings $\vl_i$. A side note: as the autoencoder is trained using our constructed, more consistent, averaged CLIP embedding, it also has an easier job in learning the reconstruction.

\subsection{Ground Truth (GT)-anchored 3D Querying}
\label{sec:query}

With the text query vector $\vq$, the standard approach is to directly compares the CLIP features $\vq$ of the query text with the language embeddings $\vl_i$ of each Gaussian, \ie
\bea\label{eq:one-step}
\hat\gM_\vq^{\tt 3D, 1step} \triangleq \{\vg_i | ~\forall i~ \texttt{Cos}(\vl_i,\vq) \geq \text{threshold}\},
\eea
where $\texttt{Cos}$ is the cosine similarity. However, this one-step approach struggles to find a single effective threshold across different language embeddings $\vl_i$ as CLIP's image and language embeddings are known to be not well calibrated~\cite{levine2023enabling}. 
To address this challenge, we propose our GT-anchored approach for querying 3D Gaussians. This involves first retrieving the ground truth and then comparing the similarity relative to the ground truth, as described in more detail below.

{\noindent\bf GT retrieval.}  Given the CLIP feature of a text query $\vq \in \sR^{512}$, we first apply a low threshold to filter out invalid prompts. We then retrieve the most similar average feature (GT for distillation) over all regions feature  
\bea\label{eq:cos_sim_supervision}
\bar\phi^*_{r} \triangleq  \argmax_{r \in \{r'| \texttt{Cos}(\bar\phi_r', \vq) \geq \text{threshold} \} } \texttt{Cos}(\bar\phi_r, \vq).
\eea
As $\bar\phi_r$ is obtained as a weighted average of CLIP image embeddings and $\vq$ comes from CLIP text embeddings, a direct comparison between them through cosine similarity is effective and does not involve a threshold. %

{\noindent\bf Relative comparison to GT.} With the retrieved GT $\bar\phi^*_r$, we compress it into lower dimension with the pretrained encoder $\texttt{E}$. Then we compute its cosine similarity with the learned language embedding $\vl_i$ for each Gaussian. Next, we threshold this similarity to retrieve the tentative set of Gaussians
\bea
\tilde\gM_\vq^{\tt 3D} = \{\vg_i | ~\forall i~ \text{Cos}(\vl_i,\texttt{E}(\bar\phi^*_r)) \geq \text{threshold}\},
\eea
where %
the querying process compares $\texttt{E}(\bar{\phi}^*_r) $ with $\vl_i$. Recall $\texttt{E}(\bar{\phi}^*_r)$ is used as supervision for training language embeddings $\vl_i$, \ie, the relevant language embeddings are trained to be similar to $\texttt{E}(\bar{\phi}^*_r)$. Therefore, any high threshold works well, which improves the queries' reliability and robustness. %

\section{Experiments}
\label{sec:experiments}
{\noindent\bf Datasets.} Following LangSplat~\cite{langsplat}, we conduct experiments on the further annotated LERF~\cite{lerf} dataset that contains a set of in-the-wild scenes and on the 3D-OVS~\cite{3dovs} dataset, which includes a collection of long-tail objects for evaluating open-vocabulary 3D querying. Additionally, we report results on the Replica~\cite{replica} dataset, which has labeled point clouds for indoor objects. We then evaluate on ten object classes for querying over eight senses in Replica.

{\noindent\bf Evaluation metrics.} As the ground-truths for LERF and 3D-OVS datasets are provided in 2D, we measure the performance of 3D querying indirectly. We render the queried Gaussians in $\hat\gM_\vq^{\tt 3D}$ to obtain a 2D mask for evaluation using 2D metrics following LangSplat~\cite{langsplat}. This includes mean Intersection over Union (mIoU $\uparrow$) and localization accuracy (Loc. Acc $\uparrow$). 
Here, mIoU is defined as the ratio of the intersecting pixels to the total number of pixels in the union of the predicted masks and ground-truth masks. 

For Loc. Acc, a query is considered correct if the center of the queried mask’s exterior bounding box falls within the bounding box of the ground-truth. We also report mIoU accuracy (mAcc$\uparrow$), a 2D metric proposed by OpenGaussian~\cite{opengaussian}, where a query is considered correct if its IoU is greater than $0.25$. For the Replica dataset, which contains 3D retrieval ground-truths, mIoU and mAcc are instead computed on the set of 3D locations of the Gaussians.

{\noindent\bf Baselines.} 
For a fair comparison, we strictly followed {\it OpenGaussian}~\cite{opengaussian} for the task of open-vocabulary 3D (point-level) querying. As they reported on the LERF dataset, we directly included their results from the paper. For 3D-OVS and Replica datasets, we use their publicly released implementation. To further benchmark the performance, we included more baseline methods \textbf{modified (m) for} direct 3D querying: {\it LangSplat-m}~\cite{langsplat}, which also trains a language 3D Gaussian. Hence, the standard query approach in~\equref{eq:one-step} can be directly used, following how OpenGaussian~\cite{opengaussian} evaluates. {\it GaussianGrouping-m}~\cite{ye2025gaussian}, which we follow their implementation for the open-vocabulary query to select group IDs, and use the corresponding Gaussian points as candidates.

{\noindent\bf Implementation Details.} To extract language features, we use the OpenCLIP ViT-B/16 model. For 2D mask segmentation, we employ the SAM ViT-H model, and for tracking masks of the same object, we utilize the SAM2-hiera-large model. Pretraining the standard 3D Gaussian Splatting takes 30,000 steps. This is followed by training the language embeddings for an additional 30,000 steps, skipping the densification stage. Experiments are conducted on an NVIDIA A100 GPU.

\subsection{Quantitative results}
{\bf\noindent %
LERF dataset.} In~\tabref{tab:lerf}, we show the results on the LERF dataset. We observe that Ours consistently outperforms LangSplat-m and, on average, is better than OpenGaussian, achieving an improvement of $+4.14$ in mIoU and $+10.66$ in mAcc. We believe this gain is significant, as retraining the language Gaussians 5 times with different seeds has a small standard deviation, \eg, on figurines the std is 0.24. We observe that LangSplat-m and GaussianGrouping-m, methods designed for 2D querying, do not generalize well to point-level querying. 

{\noindent\bf %
3D-OVS dataset.} The results for the 3D-OVS dataset is reported in~\tabref{tab:3dovs}. Our method achieves 70.58\% in mIoU and 93.33\% in Loc. Acc, significantly outperforming baseline methods. Notably, our method demonstrates a $+20.42$ gain in mIoU averaged across scenes. We observe that there exists $100\%$ in the Loc. Acc because the dataset is relatively easy with $\leq 30$ frames and $\leq 7$ objects in each scene, leading to effective segmentation and tracking from SAM2.
Next, we observed that OpenGaussian has a low Loc. Acc due to small ``floaters'' in the retrieved Gaussians. These floaters shift the center point of the exterior bounding box to an incorrect location.

\begin{figure}[t]
\centering
\includegraphics[width=0.98\linewidth]{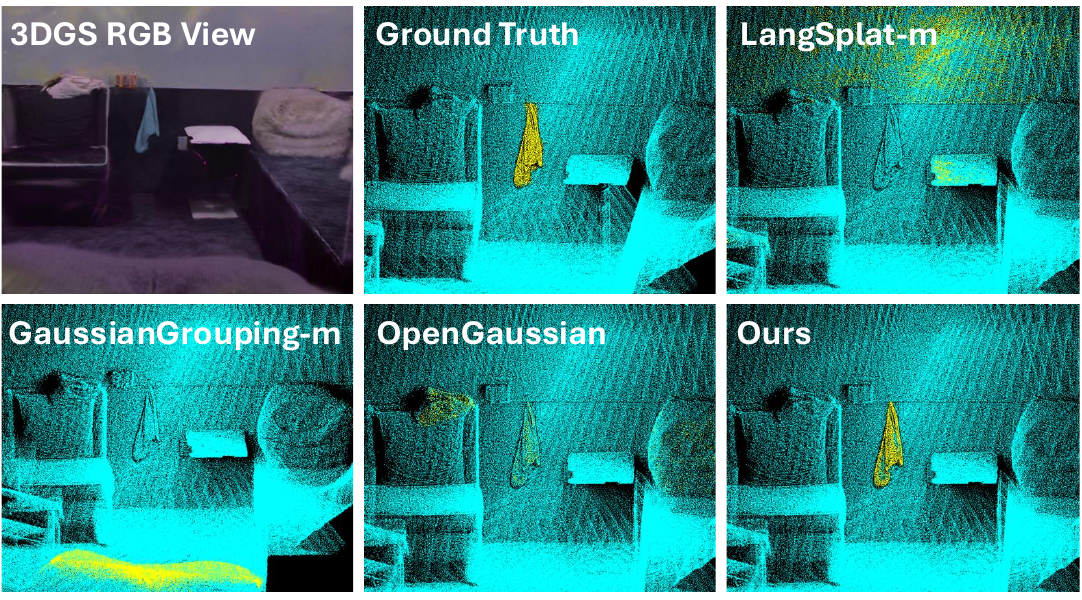}
\vspace{-.1cm}
\caption{Qualitative results on Replica dataset.
Yellow points are the queried points of \textit{``cloth''}. 
}
\vspace{-.34cm}
\label{fig:replica_qualitative}
\end{figure}

{\noindent\bf %
Replica dataset.} We show results in~\tabref{tab:replica}. We observe that our approach outperforms OpenGaussian with an average gain of $+1.74$ on mIoU and $+9.29$ on mAcc. There exists $0.00$ in GaussianGrouping-m's results because their query implementation only uses the first frame's semantics. In large scenes like Replica, the query object may not appear initially, causing an empty query.

\subsection{Qualitative results}

{\bf\noindent LERF dataset.} To further analyze the methods, we visualize the learned language embeddings and the queried Gaussians in~\figref{fig:lerf_qualitative}. We observe that baselines produce less consistent language embeddings, evident from the blurriness along object boundaries. 
Our method achieves cleaner and more fine-grained object localization in 3D space based on text queries, as illustrated in the second row for each scene in~\figref{fig:lerf_qualitative}. Note that all four methods encounter a common failure mode of empty query, \ie, no valid Gaussians are returned for a text query, resulting in zero IoU for that query.

{\noindent\bf 3D-OVS dataset.}
\figref{fig:3dovs_qualitative} shows the qualitative results on 3D-OVS dataset. Our method creates more consistent language embeddings with crisp object boundaries. Furthermore, our method's retrieval results are less noisy and preserve the complete structure of each queried object,~\ie, the lawn is accurately retrieved with query ``green lawn".

{\noindent\bf Replica dataset.} \figref{fig:replica_qualitative} visualizes the queried points for a scene in the Replica dataset given the query ``cloth''. We observe that LangSplat-m and GaussianGrouping-m failed to retrieve the correct object, and OpenGaussian only retrieves part of the cloth with noisy points from other objects. Overall, our method retrieved a more complete and clearer object, which is consistent with the quantitative result.

\begin{table}[t]
    \small
    \centering
    \setlength{\tabcolsep}{5pt}
    \caption{Ablation study on each proposed component using LERF's ``figurines'' scene. Note that combining the language features relies on leveraging tracking information and is not feasible without it. So we put "-" in the first row. The best result is achieved with all our proposed components. %
    }
    \begin{tabular}{c c c| c c c }
     \specialrule{.10em}{.01em}{.01em}
     \multicolumn{3}{c|}{Ablations} &
     \multicolumn{3}{c}{Metrics} \\
      Tracking & $\bar\phi$ Strategy & GT-anchored
     & mIoU  & mAcc & Loc. Acc \\
    \hline
    &  - &  \checkmark & 4.56 & 7.14 & 5.36 \\
    \checkmark & $\bar\phi_r$  &  & 9.09 &  12.50 & 8.93 \\  
    \checkmark & $\bar \phi_r^{\texttt{stand.}}$ &  \checkmark&  48.84   & 67.86 & 23.21   \\     
    \rowcolor{gray!10}
    \checkmark & $\bar\phi_r$  & \checkmark & \textbf{58.91}   & \textbf{82.14} & \textbf{30.36}\\
    \specialrule{.10em}{.01em}{.01em}
    \end{tabular}
    \label{tab:ablation}
\end{table}

\begin{table}[t]
    \small
    \setlength{\tabcolsep}{1.5pt}
    \centering
    \caption{Ablation study on DBSCAN and Canonical Query on scene "figurines" in LERF.}
    \begin{tabular}{c c c c| c c c }
     \specialrule{.10em}{.01em}{.01em}
     \multicolumn{4}{c|}{Ablations} &
     \multicolumn{3}{c}{Metrics} \\
      SAM2  & GT-anchored & DBSCAN & Cano. Query
     & mIoU  & mAcc & Loc. Acc \\
    \hline
    \checkmark &  & \checkmark  & \checkmark   & 15.67  & 26.79 & 14.29 \\
    \checkmark & \checkmark &   &   & 40.06  & 66.07  & 14.50 \\
    \rowcolor{gray!10}
    \checkmark &\checkmark  & \checkmark &  & \textbf{58.91}   & \textbf{82.14} & \textbf{30.36}\\
    \specialrule{.10em}{.01em}{.01em}
    \end{tabular}
    \label{tab:ablation_cluster}
    \vspace{-.05cm}
\end{table}

\subsection{Ablation study.}
We conduct ablation studies to validate the efficacy of each proposed component of our method and report the performance in~\tabref{tab:ablation} on LERF's "figurines" scene. Recall, $\bar\phi_r$ denotes the \textit{weighted} average features. As a comparison, we also tested on a standard average strategy that is the sum of features from various views divided by the number of features, denoted as $\bar \phi_r^{\texttt{stand.}}$.
As shown, mIoU increases significantly, \ie +49.89, when using the GT-anchored query. In the meantime, SAM2's masklets to get consistent features of the region also play an important role for the GT-anchored query to work. 

Overall, we observe that all proposed components are necessary to achieve the optimal performance. 
We also studied the effectiveness of our method without DBSCAN~\cite{dbscan} and evaluated the performance of canonical querying from LERF~\cite{lerf} on the task of 3D querying. The original definition of canonical query from LERF is conducted on rendered 2D images. 

To evaluate the performance of canonical query in 3D querying, we replace $\phi_{\texttt{lang}}$ by $\vl_i$ in the original equation. Formally, for each language embedding $\vl_i$ and each text query $\vq$, the relevancy score is defined as $
\min_i \frac{\exp(\vl_i \cdot \vq)}{\exp(\vl_i \cdot \vq) + \exp(\vq \cdot \phi_{canon}^i)} $
where $\phi_{canon}^i$ is the CLIP embedding of a predefined \textit{canonical} phrase chosen from \textit{``object"}, \textit{``things"}, \textit{``stuff"}, and \textit{``texture"}. The results are shown in~\tabref{tab:ablation_cluster} on LERF's ``figurines'' scene. We see that without DBSCAN, our proposed method also gains significant improvement in performance. We also see that our proposed GT-anchored query significantly outperforms the canonical query.

\section{Conclusion}\label{sec:conclusion}
We study the task of open-vocabulary 3D understanding formulated as a point-level 3D querying task. Based on LangSplat's framework, we present a tracking-based approach to provide consistent ground-truth supervision when distilling the language features.
Furthermore, we introduce a novel GT-Anchored querying pipeline to address the difficulty of choosing a consistent threshold in the baseline's single-step query. Experiments over three datasets demonstrate that our approach achieves state-of-the-art performance, with ablation studies verifying the efficacy of the proposed components.

{\small
\bibliographystyle{plainnat}
\bibliography{main}

\begin{thebibliography}{39}
\providecommand{\natexlab}[1]{#1}
\providecommand{\url}[1]{\texttt{#1}}
\expandafter\ifx\csname urlstyle\endcsname\relax
  \providecommand{\doi}[1]{doi: #1}\else
  \providecommand{\doi}{doi: \begingroup \urlstyle{rm}\Url}\fi

\bibitem[Caron et~al.(2021)Caron, Touvron, Misra, J{\'e}gou, Mairal, Bojanowski, and Joulin]{dino}
Mathilde Caron, Hugo Touvron, Ishan Misra, Herv{\'e} J{\'e}gou, Julien Mairal, Piotr Bojanowski, and Armand Joulin.
\newblock Emerging properties in self-supervised vision transformers.
\newblock In \emph{ICCV}, 2021.

\bibitem[Cen et~al.(2023)Cen, Zhou, Fang, Shen, Xie, Jiang, Zhang, Tian, et~al.]{cen2023segment}
Jiazhong Cen, Zanwei Zhou, Jiemin Fang, Wei Shen, Lingxi Xie, Dongsheng Jiang, Xiaopeng Zhang, Qi~Tian, et~al.
\newblock Segment anything in 3d with {NeRFs}.
\newblock In \emph{NeurIPS}, 2023.

\bibitem[Chen and Wang(2024)]{chen2024survey}
Guikun Chen and Wenguan Wang.
\newblock A survey on {3D Gaussian} splatting.
\newblock \emph{arXiv preprint arXiv:2401.03890}, 2024.

\bibitem[Cherti et~al.(2023)Cherti, Beaumont, Wightman, Wortsman, Ilharco, Gordon, Schuhmann, Schmidt, and Jitsev]{cherti2023reproducible}
Mehdi Cherti, Romain Beaumont, Ross Wightman, Mitchell Wortsman, Gabriel Ilharco, Cade Gordon, Christoph Schuhmann, Ludwig Schmidt, and Jenia Jitsev.
\newblock Reproducible scaling laws for contrastive language-image learning.
\newblock In \emph{CVPR}, 2023.

\bibitem[Ester et~al.(1996)Ester, Kriegel, Sander, Xu, et~al.]{dbscan}
Martin Ester, Hans-Peter Kriegel, J{\"o}rg Sander, Xiaowei Xu, et~al.
\newblock A density-based algorithm for discovering clusters in large spatial databases with noise.
\newblock In \emph{KDD}, 1996.

\bibitem[Guo et~al.(2024)Guo, Ma, Fan, Liu, and Li]{semanticgs}
Jun Guo, Xiaojian Ma, Yue Fan, Huaping Liu, and Qing Li.
\newblock Semantic {{Gaussian}s}: Open-vocabulary scene understanding with {3D {Gaussian}} splatting.
\newblock \emph{arXiv preprint arXiv:2403.15624}, 2024.

\bibitem[He et~al.(2024)He, Ding, Jiang, and Wen]{he2025segpoint}
Shuting He, Henghui Ding, Xudong Jiang, and Bihan Wen.
\newblock {SegPoint}: Segment any point cloud via large language model.
\newblock In \emph{ECCV}, 2024.

\bibitem[Jatavallabhula et~al.(2023)Jatavallabhula, Kuwajerwala, Gu, Omama, Chen, Maalouf, Li, Iyer, Saryazdi, Keetha, et~al.]{conceptfusion}
Krishna~Murthy Jatavallabhula, Alihusein Kuwajerwala, Qiao Gu, Mohd Omama, Tao Chen, Alaa Maalouf, Shuang Li, Ganesh Iyer, Soroush Saryazdi, Nikhil Keetha, et~al.
\newblock {ConceptFusion}: Open-set multimodal {3D} mapping.
\newblock \emph{RSS}, 2023.

\bibitem[Ji et~al.(2024)Ji, Wu, Fang, Cen, Yi, Liu, Tian, and Wang]{ji2024segment}
Shengxiang Ji, Guanjun Wu, Jiemin Fang, Jiazhong Cen, Taoran Yi, Wenyu Liu, Qi~Tian, and Xinggang Wang.
\newblock Segment any 4d {Gaussian}s.
\newblock \emph{arXiv preprint arXiv:2407.04504}, 2024.

\bibitem[Ji et~al.(2025)Ji, Zhu, Tang, Liu, Zhang, Tan, and Xie]{ji2025fastlgs}
Yuzhou Ji, He~Zhu, Junshu Tang, Wuyi Liu, Zhizhong Zhang, Xin Tan, and Yuan Xie.
\newblock Fastlgs: Speeding up language embedded gaussians with feature grid mapping.
\newblock In \emph{AAAI}, 2025.

\bibitem[Kerbl et~al.(2023)Kerbl, Kopanas, Leimk{\"u}hler, and Drettakis]{3dgs}
Bernhard Kerbl, Georgios Kopanas, Thomas Leimk{\"u}hler, and George Drettakis.
\newblock {3D {Gaussian}} splatting for real-time radiance field rendering.
\newblock \emph{ACM TOG}, 2023.

\bibitem[Kerr et~al.(2023)Kerr, Kim, Goldberg, Kanazawa, and Tancik]{lerf}
Justin Kerr, Chung~Min Kim, Ken Goldberg, Angjoo Kanazawa, and Matthew Tancik.
\newblock {LERF}: Language embedded radiance fields.
\newblock In \emph{ICCV}, 2023.

\bibitem[Kim et~al.(2024)Kim, Wu, Kerr, Goldberg, Tancik, and Kanazawa]{kim2024garfield}
Chung~Min Kim, Mingxuan Wu, Justin Kerr, Ken Goldberg, Matthew Tancik, and Angjoo Kanazawa.
\newblock {GARField}: Group anything with radiance fields.
\newblock In \emph{CVPR}, 2024.

\bibitem[Kirillov et~al.(2023)Kirillov, Mintun, Ravi, Mao, Rolland, Gustafson, Xiao, Whitehead, Berg, Lo, et~al.]{sam1}
Alexander Kirillov, Eric Mintun, Nikhila Ravi, Hanzi Mao, Chloe Rolland, Laura Gustafson, Tete Xiao, Spencer Whitehead, Alexander~C Berg, Wan-Yen Lo, et~al.
\newblock Segment anything.
\newblock In \emph{ICCV}, 2023.

\bibitem[LeVine et~al.(2023)LeVine, Pikus, Raja, and Gil]{levine2023enabling}
Will LeVine, Benjamin Pikus, Pranav Raja, and Fernando~Amat Gil.
\newblock Enabling calibration in the zero-shot inference of large vision-language models.
\newblock In \emph{ICLR Workshop on Trustworthy ML}, 2023.

\bibitem[Li et~al.(2022)Li, Weinberger, Belongie, Koltun, and Ranftl]{lseg}
Boyi Li, Kilian~Q Weinberger, Serge Belongie, Vladlen Koltun, and Ren{\'e} Ranftl.
\newblock Language-driven semantic segmentation.
\newblock \emph{ICLR}, 2022.

\bibitem[Liang et~al.(2024)Liang, Wang, Li, Niemeyer, Gasperini, Navab, and Tombari]{supergseg}
Siyun Liang, Sen Wang, Kunyi Li, Michael Niemeyer, Stefano Gasperini, Nassir Navab, and Federico Tombari.
\newblock {SuperGSeg}: Open-vocabulary {3D} segmentation with structured super-gaussians.
\newblock \emph{arXiv preprint arXiv:2412.10231}, 2024.

\bibitem[Liu et~al.(2023)Liu, Zhan, Zhang, Xu, Yu, El~Saddik, Theobalt, Xing, and Lu]{3dovs}
Kunhao Liu, Fangneng Zhan, Jiahui Zhang, Muyu Xu, Yingchen Yu, Abdulmotaleb El~Saddik, Christian Theobalt, Eric Xing, and Shijian Lu.
\newblock Weakly supervised 3d open-vocabulary segmentation.
\newblock In \emph{NeurIPS}, 2023.

\bibitem[Liu et~al.(2024)Liu, Kong, Cen, Chen, Zhang, Pan, Chen, and Liu]{liu2024segment}
Youquan Liu, Lingdong Kong, Jun Cen, Runnan Chen, Wenwei Zhang, Liang Pan, Kai Chen, and Ziwei Liu.
\newblock Segment any point cloud sequences by distilling vision foundation models.
\newblock In \emph{NeurIPS}, 2024.

\bibitem[Martin-Brualla et~al.(2021)Martin-Brualla, Radwan, Sajjadi, Barron, Dosovitskiy, and Duckworth]{martin2021nerf}
Ricardo Martin-Brualla, Noha Radwan, Mehdi~SM Sajjadi, Jonathan~T Barron, Alexey Dosovitskiy, and Daniel Duckworth.
\newblock Nerf in the wild: Neural radiance fields for unconstrained photo collections.
\newblock In \emph{CVPR}, 2021.

\bibitem[Mildenhall et~al.(2021)Mildenhall, Srinivasan, Tancik, Barron, Ramamoorthi, and Ng]{nerf}
Ben Mildenhall, Pratul~P Srinivasan, Matthew Tancik, Jonathan~T Barron, Ravi Ramamoorthi, and Ren Ng.
\newblock {NeRF}: Representing scenes as neural radiance fields for view synthesis.
\newblock \emph{CACM}, 2021.

\bibitem[Mohiuddin et~al.(2025)Mohiuddin, Prakhya, Collins, Liu, and Borrmann]{mohiuddin2024opensu3d}
Rafay Mohiuddin, Sai~Manoj Prakhya, Fiona Collins, Ziyuan Liu, and Andr{\'e} Borrmann.
\newblock Opensu3d: Open world 3d scene understanding using foundation models.
\newblock In \emph{ICRA}, 2025.

\bibitem[Qin et~al.(2024)Qin, Li, Zhou, Wang, and Pfister]{langsplat}
Minghan Qin, Wanhua Li, Jiawei Zhou, Haoqian Wang, and Hanspeter Pfister.
\newblock {LangSplat}: 3d language {{Gaussian}} splatting.
\newblock In \emph{CVPR}, 2024.

\bibitem[Radford et~al.(2021)Radford, Kim, Hallacy, Ramesh, Goh, Agarwal, Sastry, Askell, Mishkin, Clark, et~al.]{clip}
Alec Radford, Jong~Wook Kim, Chris Hallacy, Aditya Ramesh, Gabriel Goh, Sandhini Agarwal, Girish Sastry, Amanda Askell, Pamela Mishkin, Jack Clark, et~al.
\newblock Learning transferable visual models from natural language supervision.
\newblock In \emph{ICML}, 2021.

\bibitem[Ravi et~al.(2025)Ravi, Gabeur, Hu, Hu, Ryali, Ma, Khedr, R{\"a}dle, Rolland, Gustafson, et~al.]{sam2}
Nikhila Ravi, Valentin Gabeur, Yuan-Ting Hu, Ronghang Hu, Chaitanya Ryali, Tengyu Ma, Haitham Khedr, Roman R{\"a}dle, Chloe Rolland, Laura Gustafson, et~al.
\newblock {SAM 2}: Segment anything in images and videos.
\newblock In \emph{ICLR}, 2025.

\bibitem[Shen et~al.(2023)Shen, Yang, Yu, Wong, Kaelbling, and Isola]{shen2023distilled}
William Shen, Ge~Yang, Alan Yu, Jansen Wong, Leslie~Pack Kaelbling, and Phillip Isola.
\newblock Distilled feature fields enable few-shot language-guided manipulation.
\newblock In \emph{CORL}, 2023.

\bibitem[Siddiqui et~al.(2023)Siddiqui, Porzi, Bul{\'o}, M{\"u}ller, Nie{\ss}ner, Dai, and Kontschieder]{siddiqui2023panoptic}
Yawar Siddiqui, Lorenzo Porzi, Samuel~Rota Bul{\'o}, Norman M{\"u}ller, Matthias Nie{\ss}ner, Angela Dai, and Peter Kontschieder.
\newblock Panoptic lifting for 3d scene understanding with neural fields.
\newblock In \emph{CVPR}, 2023.

\bibitem[Straub et~al.(2019)Straub, Whelan, Ma, Chen, Wijmans, Green, Engel, Mur-Artal, Ren, Verma, et~al.]{replica}
Julian Straub, Thomas Whelan, Lingni Ma, Yufan Chen, Erik Wijmans, Simon Green, Jakob~J Engel, Raul Mur-Artal, Carl Ren, Shobhit Verma, et~al.
\newblock The replica dataset: A digital replica of indoor spaces.
\newblock \emph{arXiv preprint arXiv:1906.05797}, 2019.

\bibitem[Wu et~al.(2024)Wu, Meng, Li, Wu, Shi, Cheng, Zhao, Feng, Ding, Wang, et~al.]{opengaussian}
Yanmin Wu, Jiarui Meng, Haijie Li, Chenming Wu, Yahao Shi, Xinhua Cheng, Chen Zhao, Haocheng Feng, Errui Ding, Jingdong Wang, et~al.
\newblock {Open{Gaussian}}: Towards point-level {3D} {Gaussian}-based open vocabulary understanding.
\newblock \emph{NeurIPS}, 2024.

\bibitem[Xie et~al.(2022)Xie, Takikawa, et~al.]{xie2022neural}
Yiheng Xie, Towaki Takikawa, et~al.
\newblock Neural fields in visual computing and beyond.
\newblock In \emph{Computer Graphics Forum}, 2022.

\bibitem[Xu et~al.(2024)Xu, Chen, Zhao, Wang, Zhou, and Lu]{xu2024embodiedsam}
Xiuwei Xu, Huangxing Chen, Linqing Zhao, Ziwei Wang, Jie Zhou, and Jiwen Lu.
\newblock Embodiedsam: Online segment any 3d thing in real time.
\newblock \emph{ICLR}, 2024.

\bibitem[Yamazaki et~al.(2024)Yamazaki, Hanyu, Vo, Pham, Tran, Doretto, Nguyen, and Le]{yamazaki2024openfusion}
Kashu Yamazaki, Taisei Hanyu, Khoa Vo, Thang Pham, Minh Tran, Gianfranco Doretto, Anh Nguyen, and Ngan Le.
\newblock Open-fusion: Real-time open-vocabulary 3d mapping and queryable scene representation.
\newblock In \emph{ICRA}, 2024.

\bibitem[Ye et~al.(2023)Ye, Wang, and Wang]{ye2023featurenerf}
Jianglong Ye, Naiyan Wang, and Xiaolong Wang.
\newblock Featurenerf: Learning generalizable nerfs by distilling foundation models.
\newblock In \emph{ICCV}, 2023.

\bibitem[Ye et~al.(2024)Ye, Danelljan, Yu, and Ke]{ye2025gaussian}
Mingqiao Ye, Martin Danelljan, Fisher Yu, and Lei Ke.
\newblock {Gaussian} grouping: Segment and edit anything in 3d scenes.
\newblock In \emph{ECCV}, 2024.

\bibitem[Yu et~al.(2024)Yu, Chen, Huang, Sattler, and Geiger]{yu2024mip}
Zehao Yu, Anpei Chen, Binbin Huang, Torsten Sattler, and Andreas Geiger.
\newblock {Mip-Splatting}: Alias-free {3D Gaussian} splatting.
\newblock In \emph{CVPR}, 2024.

\bibitem[Zhi et~al.(2021)Zhi, Laidlow, Leutenegger, and Davison]{zhi2021place}
Shuaifeng Zhi, Tristan Laidlow, Stefan Leutenegger, and Andrew~J Davison.
\newblock In-place scene labelling and understanding with implicit scene representation.
\newblock In \emph{ICCV}, 2021.

\bibitem[Zhou et~al.(2022)Zhou, Loy, and Dai]{zhou2022extract}
Chong Zhou, Chen~Change Loy, and Bo~Dai.
\newblock Extract free dense labels from {CLIP}.
\newblock In \emph{ECCV}, 2022.

\bibitem[Zhou et~al.(2024)Zhou, Chang, et~al.]{feature3dgs}
Shijie Zhou, Haoran Chang, et~al.
\newblock Feature {3DGS}: Supercharging 3d {{Gaussian}} splatting to enable distilled feature fields.
\newblock In \emph{CVPR}, 2024.

\bibitem[Zwicker et~al.(2001)Zwicker, Pfister, Van~Baar, and Gross]{zwicker2001ewa}
Matthias Zwicker, Hanspeter Pfister, Jeroen Van~Baar, and Markus Gross.
\newblock {EWA} volume splatting.
\newblock In \emph{IEEE VIS}, 2001.

\end{thebibliography}
}

\end{document}